\documentclass[10pt,twocolumn,letterpaper]{article}
\usepackage[pagenumbers]{cvpr}

\usepackage{amsmath,amsfonts,bm}

\def\eqref#1{equation~\ref{#1}}

\def\1{\bm{1}}

\DeclareMathAlphabet{\mathsfit}{\encodingdefault}{\sfdefault}{m}{sl}
\SetMathAlphabet{\mathsfit}{bold}{\encodingdefault}{\sfdefault}{bx}{n}

\usepackage{placeins}
\usepackage{tabularx}
\usepackage{makecell}
\usepackage{array}
\usepackage{microtype}
\definecolor{cvprblue}{rgb}{0.21,0.49,0.74}
\usepackage[pagebackref,breaklinks,colorlinks,allcolors=cvprblue]{hyperref}

\newcommand{\bestresult}[1]{\textbf{#1}}
\newcommand{\secondresult}[1]{\underline{#1}}

\DeclareRobustCommand{\method}{\textup{\textsc{RAVEL}}}
\newcommand{\actions}{\mathbf{A}}

\newcommand{\context}{\mathbf{c}}

\def\confName{CVPR}
\def\confYear{2027}
\title{\method{}: Asynchronous Rolling Inference for Flow-Based Vision–Language–Action Models}
\author{
\resizebox{\dimexpr\textwidth-2\tabcolsep-4pt\relax}{!}{\textbf{Yuhan Chen$^{1}$, Ke Yu$^{1}$, Pengfei Liu$^{2}$, Shuxun Wang$^{1}$, Yi Yang$^{1}$, Linchao Zhu$^{1}$}} \\
$^{1}$Zhejiang University \\
$^{2}$Nanyang Technological University
}
\begin{document}
\twocolumn[{%
  \begin{@twocolumnfalse}
    \maketitle
  \end{@twocolumnfalse}
  \noindent\begin{minipage}{\textwidth}
    \centering
    \captionsetup{type=figure}
    \includegraphics[width=\linewidth]{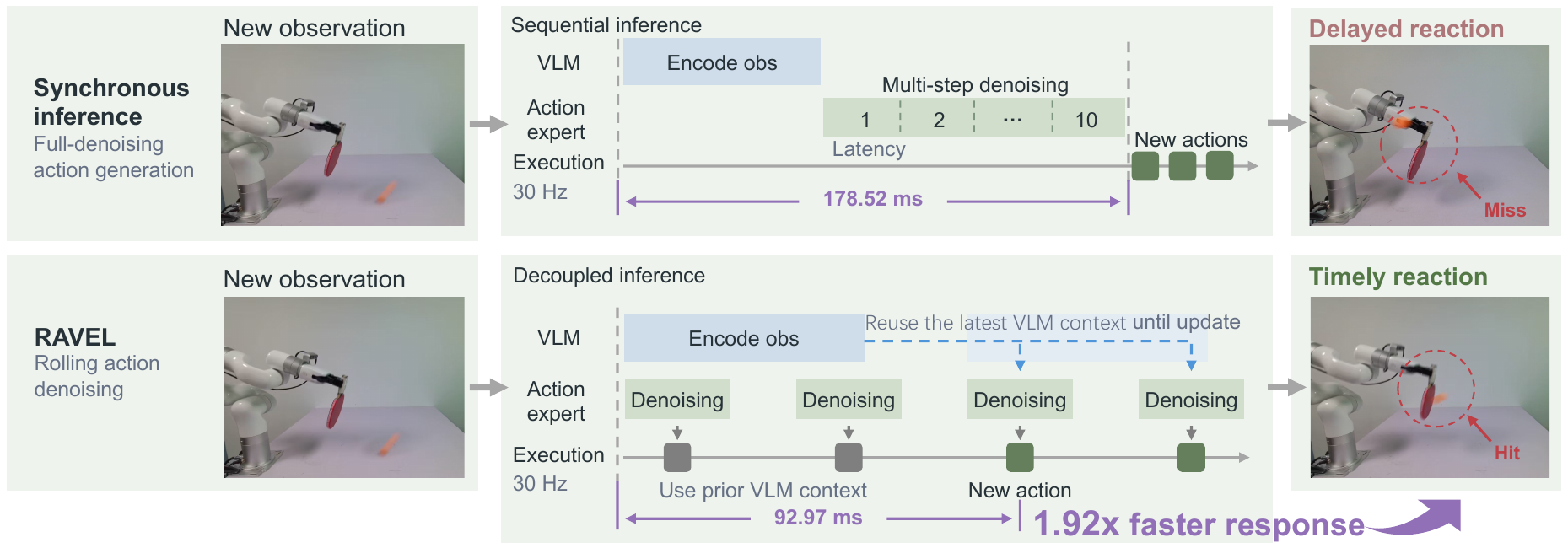}
    \caption{\textbf{Responsive control with \method{}.} Synchronous inference waits for VLM encoding and multi-step action denoising before executing new actions. \method{} runs VLM context updates in parallel with rolling action denoising and execution, enabling faster responses to new observations.}
    \label{fig:teaser}
  \end{minipage}\par
}]

\begin{abstract}
Flow-based vision--language--action (VLA) models are highly effective for generalist robot manipulation, yet their reliance on computationally expensive VLM encoding and multi-step iterative action generation imposes a significant latency bottleneck. The resulting inference latency makes it difficult for robots to respond quickly, especially in dynamic environments.
We address this limitation with \textbf{RAVEL} (\textbf{R}olling \textbf{A}synchronous \textbf{V}LA \textbf{E}nabling \textbf{L}ow-Latency Control), an asynchronous inference framework that addresses the computational bottlenecks of both the VLM backbone and the action expert.
To reduce the delay from multi-step action denoising, \textbf{RAVEL} allows near-term actions to be executed after a single additional denoising step by carrying partially denoised future actions forward in a rolling buffer.
To avoid blocking on slow VLM encoding, \textbf{RAVEL} decouples VLM encoding from rolling action generation, allowing the action expert to operate continuously using the latest available VLM context, while a lightweight Fast Observation Pathway (FOP) directly conditions the action expert on current observations.
Across simulated and real-world manipulation tasks, \textbf{RAVEL} consistently achieves substantially lower response latency while maintaining the task capability of the underlying VLA, enabling high-frequency and responsive closed-loop control.
\end{abstract}

\section{Introduction}
\label{sec:introduction}

Vision--language--action (VLA) models build on pretrained vision--language models (VLMs) to learn generalist robot policies~\citep{brohan2023rt2,kim2024openvla,team2025gemini}.
A common strategy in robot learning is to predict a sequence of future actions, or an \emph{action chunk}~\citep{zhao2023act,chi2023diffusionpolicy}.
Recent VLAs such as $\pi_{0.5}$ and X-VLA generate these action chunks using flow matching~\citep{black2025pi05,zheng2025xvla,bjorck2025grootn1}.
We refer to this family of models as \emph{flow-based VLAs}.
Given an observation and the current robot state, the model first encodes the observation with the VLM and then uses an action expert to generate an action chunk through multiple denoising steps.
Unfortunately, the sequential VLM encoding and multi-step action generation result in high inference latency~\citep{ma2025running}. Under conventional synchronous inference, the robot may pause between action chunks while waiting for inference to finish.
Executing multiple actions from a fixed chunk also limits how frequently new observations can influence control.

To address these limitations, asynchronous methods run inference while the robot executes the current action chunk, reducing pauses between action chunks~\citep{shukor2025smolvla,zhao2025vlarail}.
However, under naive asynchronous execution, a newly generated chunk may no longer match the robot state when execution begins, leading to discontinuities between chunks.
Methods such as RTC improve continuity across chunk transitions~\citep{black2025rtc,tang2025vlash,liu2026legato,black2025trainingrtc}, but each new action update still requires VLM encoding followed by iterative denoising.
Recent approaches reduce the iterative computation required before near-term actions become executable~\citep{du2026cfvla,jiang2025streamingflow}, using strategies such as horizon-aware noise schedules or rolling denoising~\citep{ruhe2024rolling,chen2024diffusionforcing,lu2026faster,hoeg2024sdp,jung2025rdp}.
Reducing denoising computation alone, however, does not eliminate the delay introduced by VLM encoding.
Responsive control therefore requires both making near-term actions executable sooner and allowing action generation to proceed without waiting for each VLM update.

To this end, we introduce \method{}, an asynchronous inference framework for flow-based VLAs (Figure~\ref{fig:teaser}).
To reduce denoising latency, we progressively denoise future actions across control steps using a rolling action buffer.
At each control step, the action expert denoises the buffer once, releases the first action, and appends a new noisy action while retaining the remaining actions for further denoising.
The next action therefore requires only one final denoising update before execution.
Meanwhile, we decouple VLM encoding from action generation to prevent expensive VLM forward passes from blocking this rolling process.
VLM context is updated asynchronously, while the action expert continues denoising the buffer using the latest available VLM context.
To generate actions for the current control step from delayed VLM context, we condition the action expert on the temporal offset between the source observation and the current control step.
This is enabled by streaming-aligned training, which trains the expert to refine partially denoised action buffers using temporally shifted observation--action pairs and their corresponding offsets.
We further introduce a lightweight Fast Observation Pathway (FOP) that directly supplies current visual and proprioceptive features to the action expert, allowing upcoming actions to respond to recent observations without waiting for VLM encoding to finish.

Our evaluation spans both static and dynamic manipulation tasks in simulation and the real world.
Results on LIBERO~\citep{liu2023libero} and RoboTwin~2.0~\citep{chen2025robotwin2} show that adapting the base VLAs to our framework maintains comparable task performance. Furthermore, evaluations on the Dynamic Object Manipulation benchmark~\citep{xie2026dynamicvla} show improved task success relative to the evaluated baselines.
Separate latency measurements show reduced observation-to-command response times.
Finally, real-world experiments on table-tennis return, conveyor-belt grasping, and tennis-ball grasping demonstrate our method's ability to respond rapidly and produce smooth motion.

Our main contributions are as follows:
\begin{itemize}
    \item We propose \method{}, an asynchronous inference framework for flow-based VLAs that reduces observation-to-action latency through rolling denoising. It enables continuous action refinement during VLM encoding by leveraging the latest available VLM context and fresh observations through a lightweight Fast Observation Pathway (FOP).
    \item We introduce streaming-aligned training to enable action generation from delayed VLM context. By training on temporally shifted observation--action pairs and conditioning on the corresponding temporal offset, the action expert learns to generate actions aligned with the current control step.
    \item We demonstrate that \method{} preserves the manipulation capabilities of pretrained VLAs while substantially reducing response latency, improving success rates in dynamic simulation, and enabling faster responses and smoother motion in real-world manipulation.
\end{itemize}

\section{Related Work}
\label{sec:related}

\paragraph{Flow-based VLA and action chunking.}
Recent VLAs~\citep{li2024cogact,shao2025vlasurvey} often combine a pretrained vision-language model (VLM) with a flow-matching action expert~\citep{lipman2023flow}, as in $\pi_{0.5}$ and X-VLA~\citep{black2025pi05,zheng2025xvla}, among other recent models~\citep{black2024pi0,bjorck2025grootn1,shukor2025smolvla,wang2026qwenvla,wu2026lingbotvla}. Since running the full VLA policy is expensive, these methods typically predict a chunk of future actions~\citep{zhao2023act,liu2024rdt} at each inference step. However, action chunking still requires VLM encoding followed by multi-step action denoising before the first action becomes available. Our work addresses this latency bottleneck in flow-based VLAs.

\paragraph{Real-time and asynchronous action generation.}
Asynchronous action-chunking methods overlap inference with execution while improving alignment between predicted actions and their execution times~\citep{black2025rtc,black2025trainingrtc,tang2025vlash,xie2026dynamicvla,wang2026remac}.
To reduce action-generation latency, RDP and RNR-DP progressively refine pending actions, while Streaming Flow Policy generates actions during sampling by integrating a flow directly in action space~\citep{hoeg2024sdp,jung2025rdp,chen2025rnrdp,chen2025falcon,lu2026faster,jiang2025streamingflow}.
Another line of work addresses the delay of costly VLM encoding by decoupling slow perception from fast control or incorporating current observations.
For example, HiRT adopts a hierarchical policy, while Reflex enables cache-aware asynchronous perception and action generation~\citep{zhang2025hirt,guo2026reflex,sendai2025a2c2,ma2025running}.
Building on these directions, \method{} adapts pretrained flow-based VLAs to rolling action refinement with asynchronously updated VLM context, while incorporating the latest observations.
By explicitly accounting for the temporal gap between the source observation and the current control step, \method{} maintains temporal alignment throughout continuous action refinement.

\begin{figure*}[!t]
\centering
\includegraphics[width=\linewidth]{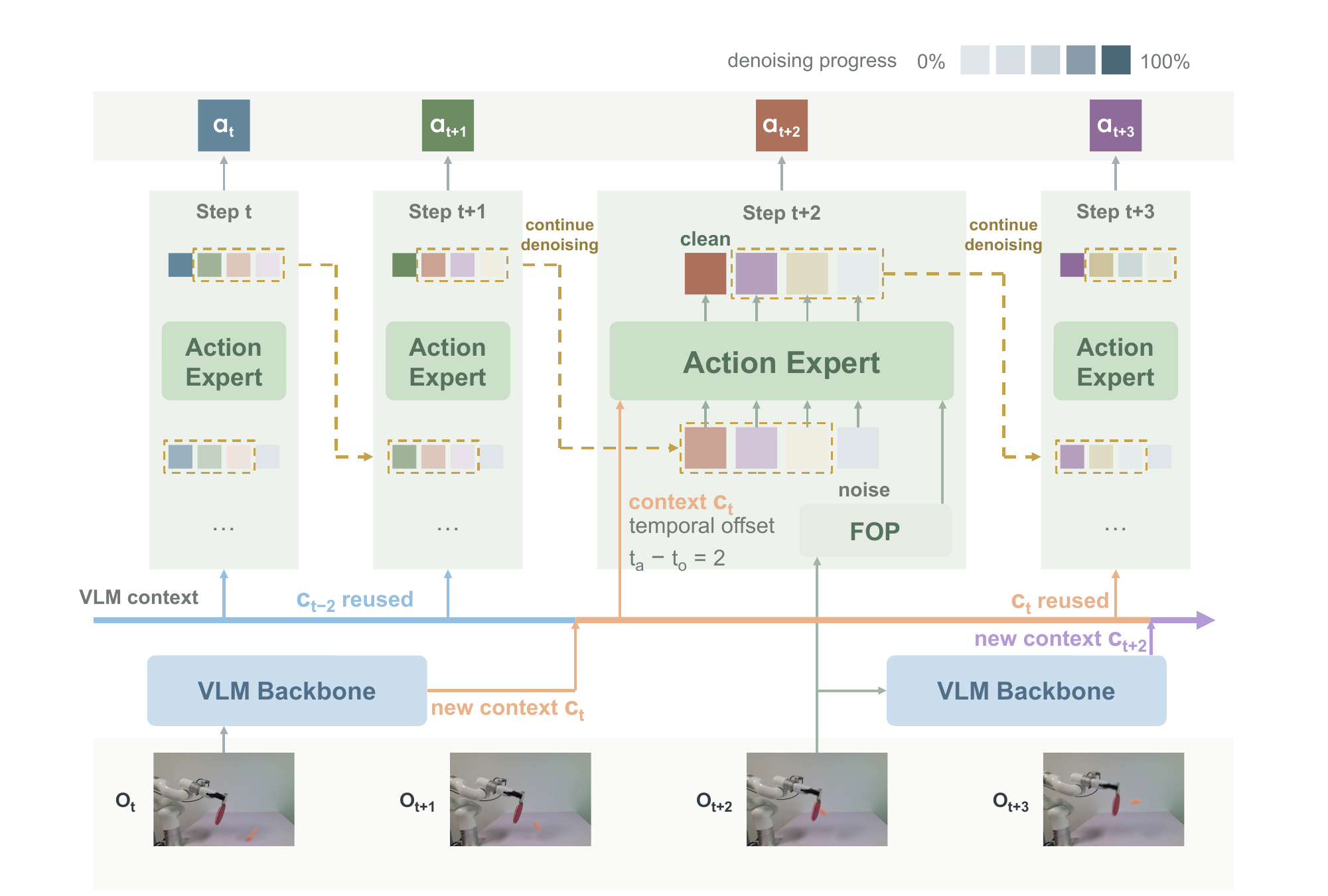}
\caption{Overview of \method{}. The VLM begins encoding observation $o_t$ at time $t$ and produces VLM context $\context_t$ between $t+1$ and $t+2$, while the action expert continues using the previously available context. At $t+2$, the expert uses $\context_t$ with a temporal offset of two control steps. Meanwhile, FOP directly conditions the action expert on the current observation $o_{t+2}$. Conditioned on these inputs, the expert advances the rolling buffer by one denoising step and releases the first action. The remaining actions are retained, and a new noisy action is appended.}
\label{fig:framework}
\end{figure*}

\section{Method}
\label{sec:method}

\subsection{Offset-Aware Rolling Flow Denoising}
\label{sec:sliding-overview}

\method{} maintains a rolling buffer of future actions, where actions scheduled
for later execution have higher noise levels (Figure~\ref{fig:framework}).
Let $K$ denote the buffer length. At control timestep $t_a$, the buffer stores
$K$ future actions:
\begin{equation}
\actions_{t_a}
=
[a_{t_a+1}^{\tau_1};\ldots;a_{t_a+k}^{\tau_k};\ldots;
a_{t_a+K}^{\tau_K}],
\label{eq:rolling-buffer}
\end{equation}
where the $k$-th slot corresponds to the action executed at $t_a+k$, and
$\tau_k=1-k/K$ denotes its flow timestep.

To update the actions in the buffer, the action expert uses a representation
produced by the VLM backbone, such as layer-wise key--value
caches~\citep{black2024pi0,black2025pi05} or VLM
features~\citep{bjorck2025grootn1,shukor2025smolvla,zheng2025xvla}, which we
refer to as the \emph{VLM context}. Given an observation $o_{t_o}$,
proprioceptive state $s_{t_o}$, and language instruction $\ell$, we denote the
VLM context as
\begin{equation}
    \context_{t_o}
    =
    E(o_{t_o}, s_{t_o}, \ell),
\label{eq:vlm-context}
\end{equation}
At control timestep $t_a$, the action expert uses a VLM context computed from an earlier observation at $t_o$. 
To account for this temporal mismatch, we introduce a normalized \emph{temporal offset} $\delta_p=p/K$, where $p=t_a-t_o$, as an additional conditioning signal for the action expert.
Given the buffer, VLM context, and temporal offset, one action-expert forward pass advances every slot by $\Delta\tau=1/K$:
\begin{equation}
a_{t_a+k}^{\tau_{k-1}}
=
a_{t_a+k}^{\tau_k}
+
\Delta\tau
\left[
\mathbf v_\theta(
\actions_{t_a},
\boldsymbol{\tau},
\context_{t_o},
\delta_p)
\right]_{[k]},
\label{eq:rolling-update}
\end{equation}
where $\mathbf v_\theta$ is the flow-based action expert,
$[\cdot]_{[k]}$ selects the velocity for slot $k$, and $\tau_0=1$
denotes the clean endpoint. 

After the update, the first action reaches $\tau_0$ and is released for execution. The remaining $K-1$ actions are kept in the buffer, and a new action initialized from Gaussian noise, $a_{t_a+K+1}^{\tau_K} \sim \mathcal{N}(0,I)$, is added as the last slot. The buffer therefore again contains $K$ future actions with flow timesteps $\tau_1,\ldots,\tau_K$ at the next control timestep. In this way, each control timestep requires only one action-expert forward pass, while each action is refined over $K$ consecutive updates before execution.

\subsection{Asynchronous Context Updates and Fast Observation Pathway}
\label{sec:async-inference}
Computing the VLM context is much slower than one action denoising step. We therefore decouple the VLM from the action expert, allowing context updates and action denoising to run asynchronously~\citep{guo2026reflex}.
At control timestep $t_a$, the action expert uses the latest available context
$\context_{t_o}$ together with its temporal offset $\delta_p$. The same context
is reused across several rolling buffer updates until a new one becomes available.

Since the VLM context is computed from an earlier observation, we introduce a Fast Observation Pathway (FOP) to provide the action expert with more recent visual and proprioceptive information. FOP uses a lightweight image encoder together with a projection of the current proprioceptive state:
\begin{equation}
z_{t_a}=F(o_{t_a},s_{t_a}).
\label{eq:fop}
\end{equation}

The resulting tokens are injected into selected action-expert layers through
gated cross-attention adapters. 
The VLM context encodes semantic information from the instruction and observation at $t_o$, while FOP incorporates the current observation to refine the action prediction.
This allows upcoming actions to respond to
new observations without waiting for the next VLM context update.

\subsection{Streaming-Aligned Training}
\label{sec:training}

Let $\hat a_t$ denote the target action at timestep $t$. Given a training
trajectory, we sample an observation timestep $t_o$ and a temporal offset
$p$, with $t_a=t_o+p$. The VLM context $\context_{t_o}$ is computed from
the observation at $t_o$, with the $K$-step action horizon
$\hat{\actions}_{t_a}=[\hat a_{t_a+1};\ldots;\hat a_{t_a+K}]$
as the training target.
By varying $p$ and providing the normalized
offset $\delta_p=p/K$ to the action expert, we train it to generate
actions for different control timesteps from a reused VLM context.

We construct the training buffer $\actions_{t_a}$ using the same
flow-timestep profile $\tau_k=1-k/K$ as in
Section~\ref{sec:sliding-overview}.
For each slot $k$, we sample $\epsilon_{t_a+k}\sim\mathcal{N}(0,I)$ and construct
\begin{equation}
a_{t_a+k}^{\tau_k}
=
(1-\tau_k)\epsilon_{t_a+k}
+
\tau_k\hat a_{t_a+k}.
\end{equation}
This yields the same position-dependent noise profile as the rolling
buffer used during inference, with later actions receiving higher noise levels.

The flow-matching target for each slot $k$ is
$\hat a_{t_a+k}-\epsilon_{t_a+k}$. We train the action expert to predict
these slot-wise velocities conditioned on the training buffer, VLM context,
and temporal offset:
\begin{equation}
\begin{aligned}
\mathcal L_{\mathrm{train}}
&=\mathbb E_{t_o,\,p,\,k}\Bigl[
\Bigl\|
\left[\mathbf v_\theta(\actions_{t_a},\boldsymbol{\tau},
\context_{t_o},\delta_p)\right]_{[k]}\\
&\hspace{5em}-(\hat a_{t_a+k}-\epsilon_{t_a+k})
\Bigr\|_2^2\Bigr].
\end{aligned}
\label{eq:training-objective}
\end{equation}
When FOP is enabled, the FOP features $z_{t_a}$ defined in
Eq.~(\ref{eq:fop}) are additionally provided to the action expert.

\section{Experiments}
\label{sec:experiments}

Our evaluation spans both static and dynamic manipulation tasks in simulation and the real world, covering manipulation capability, responsiveness, and inference efficiency.
We first use synchronous evaluations on LIBERO~\citep{liu2023libero} and RoboTwin~2.0
~\citep{chen2025robotwin2} to test whether \method{} preserves the manipulation
capability of the underlying VLAs. We then evaluate \method{} on the Dynamic
Object Manipulation (DOM) benchmark~\citep{xie2026dynamicvla}, where the
environment continues to evolve during inference, to assess responsiveness
under continuous motion. 
Finally, we deploy \method{} on real-world
table-tennis return, conveyor-belt grasping, and tennis-ball grasping to evaluate its effectiveness
under real-world timing and execution constraints.
We further analyze latency and ablate the key components of \method{}.

\subsection{Synchronous Simulation}
\label{sec:sync-non-regression}

\paragraph{Experimental Setup.}
We evaluate whether adapting pretrained flow-based VLAs to \method{} preserves
their manipulation capability. We test $\pi_{0.5}$ and X-VLA on
LIBERO~\citep{liu2023libero} and RoboTwin~2.0~\citep{chen2025robotwin2},
covering single-arm tabletop and bimanual manipulation tasks. Both benchmarks
use synchronous simulation: the environment pauses during policy inference and
resumes only after the predicted actions are returned. Inference latency
therefore does not affect the rollout, allowing us to isolate manipulation
capability from the responsiveness benefits of asynchronous execution.

For each evaluation setting, we compare the baseline with a \method{} variant
initialized from the same publicly released benchmark-specific checkpoint.
We use \method{} without FOP to assess whether it preserves the baseline
performance under the same generation budget.
\method{} is trained with streaming-aligned training (Section~\ref{sec:training}) while keeping the vision--language
backbone frozen. On both LIBERO and RoboTwin~2.0, we adapt each checkpoint
using the same demonstrations as in its original benchmark fine-tuning.
Both methods retain the native action representation and
horizon and use the same generation budget of one VLM prefill and ten
action-expert forward passes. The baseline performs ten full-horizon denoising
steps, whereas \method{} performs ten rolling updates. All methods use
the same task configurations, environment seeds, and language instructions.

We additionally measure first-action latency from receiving an observation to
making the first predicted action available for execution. Measurements are
conducted on one RTX~4090, and we report the median latency. Further training, evaluation, and buffer-handling
details are provided in Appendix~A.1.

\begin{figure}[t]
\centering
    \includegraphics[width=\linewidth]{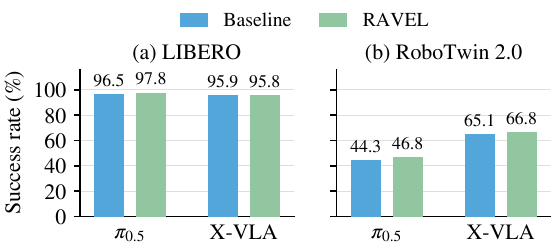}
    \caption{Synchronous evaluation of the baseline and \method{} under matched
    generation budgets. Results aggregate 2,000 LIBERO trials and 5,000
    RoboTwin~2.0 trials.}
    \label{fig:sync-success}
\end{figure}

\begin{table}[t]
\centering
    \caption{Median first-action latency for X-VLA and $\pi_{0.5}$
    using PyTorch inference on one RTX~4090, measured from observation receipt
    to the first predicted action becoming available for execution.}
    \label{tab:speed}
    \begin{tabular*}{\linewidth}{@{\extracolsep{\fill}}l@{}c@{}c@{}c@{}}
        \toprule
        Model & \makecell{Full\\denoising\\(ms)$\downarrow$}
              & \makecell{Rolling\\inference\\(ms)$\downarrow$}
              & Speedup \\
        \midrule
        X-VLA & 142.58 & \textbf{54.97} & 2.59$\times$ \\
        $\pi_{0.5}$ & 272.86 & \textbf{93.45} & 2.92$\times$ \\
        \bottomrule
    \end{tabular*}
\end{table}

\textbf{Results.}\quad
As shown in Figure~\ref{fig:sync-success}, \method{} with rolling inference
achieves comparable or slightly higher success rates than the original policies
with full-chunk denoising on LIBERO and RoboTwin~2.0 under the same generation
budget. These results show that both pretrained VLAs retain their manipulation
capabilities when adapted to rolling inference through streaming-aligned
training.

\method{} also substantially reduces first-action latency. As shown in
Table~\ref{tab:speed}, it achieves a 2.59--2.92$\times$ speedup on one
RTX~4090. This speedup is expected, as rolling inference requires only a single
additional action-expert forward pass to release each new action.

\subsection{Asynchronous Simulation}
\label{sec:dom}

\paragraph{Experimental Setup.}
We evaluate on the Dynamic Object Manipulation (DOM) benchmark introduced
with DynamicVLA~\citep{xie2026dynamicvla}, which uses moving-object
grasp-and-place tasks to assess a model's ability to interact with moving
objects, perceive dynamic scenes, and generalize across variations in
appearance and motion. 
Unlike the synchronous benchmarks above, the simulator continues to advance during policy inference. Objects may move or change direction before the predicted actions are executed. 
This setting exposes the effect of inference latency on task execution. We use DOM to evaluate the performance of \method{} in dynamic environments.

All methods are initialized from the released DynamicVLA checkpoint. We evaluate
\method{} with and without the Fast Observation Pathway (FOP), and compare
against synchronous DynamicVLA, Continuous Inference with Latent-aware
Action Streaming (CI \& LAAS)~\citep{xie2026dynamicvla},
VLASH~\citep{tang2025vlash}, and FASTER~\citep{lu2026faster}. Each method is evaluated across 9 dimensions,
with 10 scenes per dimension and 20 trials per scene, yielding 1,800 episodes
in total. We report the overall success rate and the mean task completion
time over episodes successful for all six methods, where completion time reflects environment
execution time rather than model inference latency. All experiments are run
locally on two NVIDIA RTX~4090 GPUs, with simulation and policy inference
running on separate GPUs. Additional implementation details and measured
inference latencies are provided in Appendix~A.2.

\begin{table*}[t]
    \centering
    \captionof{table}{Success rates (\%) and mean completion time over common-success
    episodes on DOM. All results are obtained from local evaluations using
    two NVIDIA RTX~4090 GPUs, with simulation and policy inference on separate
    GPUs. \bestresult{Bold} and \secondresult{underlined} values indicate the
    best and second-best results, respectively.}
    \label{tab:dom-main}
    \begin{center}
    \begin{tabularx}{\linewidth}{@{}l@{\enspace}*{11}{>{\centering\arraybackslash}X@{}}}
        \toprule
        & \multicolumn{3}{c}{Interaction}
        & \multicolumn{3}{c}{Perception}
        & \multicolumn{3}{c}{Generalization} & \multicolumn{2}{c}{Average} \\
        \cmidrule(lr){2-4}\cmidrule(lr){5-7}\cmidrule(lr){8-10}\cmidrule(lr){11-12}
        Method & CR & DA & LS & VU & SR & MP & VG & MG & DR & SR $\uparrow$ & Time $\downarrow$ \\
        \midrule
        DynamicVLA
        & 25.5 & 22.0 & 13.5 & 38.5 & 20.0 & 23.0 & 22.0 & 53.5 & 16.5 & 26.06 & 7.85 \\
        \quad + CI \& LAAS
        & 46.0 & 23.5 & 28.5 & 38.5 & 39.0 & 23.0 & 39.5 & 58.5 & \secondresult{23.0} & 35.50 & 6.58 \\
        \quad + VLASH
        & 18.5 & 16.0 & 9.5 & 13.5 & 13.5 & 17.0 & 23.5 & 40.0 & 17.5 & 18.78 & 7.13 \\
        \quad + FASTER
        & 37.0 & 19.0 & 14.5 & 20.5 & 22.5 & 24.0 & 19.0 & 45.0 & 12.5 & 23.78 & 7.11 \\
        \midrule
        \quad + \method{}
        & \secondresult{49.0}
        & \secondresult{29.0}
        & \secondresult{38.0}
        & \secondresult{48.0}
        & \secondresult{46.0}
        & \secondresult{38.5}
        & \bestresult{50.0}
        & \secondresult{70.5}
        & \bestresult{27.5}
        & \secondresult{44.06} & \bestresult{6.04} \\
        \quad + \method{} w/ FOP
        & \bestresult{62.5}
        & \bestresult{32.5}
        & \bestresult{40.5}
        & \bestresult{49.0}
        & \bestresult{49.0}
        & \bestresult{43.0}
        & \secondresult{49.0}
        & \bestresult{76.5}
        & 20.5
        & \bestresult{46.94} & \secondresult{6.19} \\
        \bottomrule
    \end{tabularx}
    \end{center}
    \begin{minipage}{\linewidth}
        CR: Closed-loop Reactivity; DA: Dynamic Adaptation;
        LS: Long-horizon Sequencing; VU: Visual Understanding;
        SR (Perception): Spatial Reasoning; MP: Motion Perception;
        VG: Visual Generalization; MG: Motion Generalization;
        DR: Disturbance Robustness; SR (Average): Success Rate.
        \par\smallskip
        Time: Mean environment steps divided by 25\,Hz, in seconds, over
        the same 36 scene--trial pairs successful for all six methods.
    \end{minipage}
\end{table*}

\paragraph{Results.}
Table~\ref{tab:dom-main} summarizes the results on DOM. \method{} achieves
an overall success rate of 44.06\%, outperforming CI \& LAAS by 8.56 percentage
points and improving performance across all nine evaluation dimensions.
Adding FOP further increases the overall success rate to 46.94\%.
In particular, \method{} with FOP reaches 62.5\% in Closed-loop Reactivity
and 43.0\% in Motion Perception, highlighting the benefit of incorporating
recent observations for tasks that require timely responses to changing
object states. \method{} also completes tasks in fewer execution steps,
achieving the shortest mean completion time of 6.04\,s, a 23.0\%
reduction relative to DynamicVLA. Together with the higher overall success
rate, this result suggests that \method{}'s latency reduction translates
into improved task execution efficiency in dynamic manipulation.
Overall, these results show that reducing action-generation latency
substantially improves performance in dynamic manipulation, while incorporating
recent visual feedback through FOP further improves adaptation to changing
object states.
We also observe that several existing methods do not improve
over the DynamicVLA baseline under our local evaluation setting. Further
implementation details and discussion are provided in
Appendix~A.2.

\subsection{Real-World Evaluation}
\label{sec:real-world}

\paragraph{Experimental Setup.}
We evaluate \method{} on three real-world tasks using an xArm7 robot:
table-tennis return, conveyor-belt grasping, and tennis-ball grasping.
All real-world policy inference runs on an NVIDIA RTX 5090 GPU.
We evaluate each method over 20 trials per task.

\begin{figure*}[!t]
    \centering
    \includegraphics[width=\linewidth]{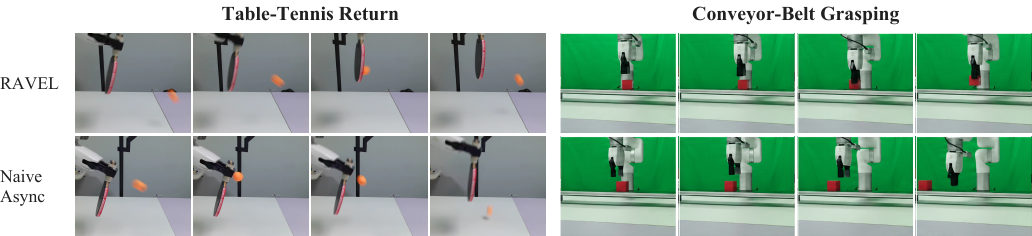}
    \caption{\textbf{Real-world rollouts on two dynamic tasks.}
    \textbf{Left:} Table-tennis return. \method{} swings in time to return
    the incoming ball. Naive Async only begins preparing to swing
    after the incoming ball has already hit the paddle handle.
    \textbf{Right:} Conveyor-belt grasping. \method{} grasps and lifts
    the moving cube, while Naive Async misses the grasp as the cube
    passes the gripper.}
    \label{fig:real-staff-async-demo}
\end{figure*}

For table-tennis return, the robot must position the paddle to intercept
the incoming ball and time its swing to return the ball. For conveyor-belt
grasping, the robot must align the gripper with the moving cube and close
it before the cube passes the grasp point. We compare \method{} with
synchronous inference (Sync), naive
asynchronous inference (Naive Async), RTC~\citep{black2025rtc}, and
VLASH~\citep{tang2025vlash}. All policies use the same teleoperated
training data for each task, comprising 299 trajectories for
table-tennis return and 272 for conveyor-belt grasping. We select
appropriate inference parameters for each method. We evaluate task
success rates and observation age, which measures the elapsed time between the policy's
receipt of the source image used to generate an action and the dispatch
of that action to the robot. We report the median (P50) and 95th
percentile (P95) observation age. Representative rollouts for both
dynamic tasks are shown in
Figure~\ref{fig:real-staff-async-demo}.

For tennis-ball grasping, a trial succeeds when the robot grasps the ball
and successfully places it in the basket. We compare \method{}
with Sync, Naive Async, and RTC. In addition to grasping success rate, we
evaluate motion quality to examine whether the different inference schemes
affect the smoothness and stability of robot execution. Specifically, we
report the high-frequency residual power $E_{\mathrm{HF}}$, inspired by PSD-based trajectory analysis~\citep{sojib2026efficientmetric}, to quantify high-frequency oscillations in
the TCP trajectory, and TCP jerk RMS to measure abrupt changes in motion.
Lower $E_{\mathrm{HF}}$ indicates less high-frequency trajectory variation,
while lower jerk RMS indicates smoother changes in acceleration. Together,
these metrics allow us to evaluate not only whether the robot completes the
grasp, but also how smoothly and stably the motion is executed. Detailed
metric definitions and measurement protocols are provided in
Appendix~\ref{app:real-metrics}.

\begin{table}[t]
    \centering
    \caption{\textbf{Real-world table-tennis return and conveyor-belt grasping.}
    Observation age is reported as the median (P50) and 95th percentile (P95).
    Bold indicates the best result in each column.}
    \label{tab:real-world}
    \small
    \setlength{\tabcolsep}{3pt}
    \begin{tabular*}{\linewidth}{@{\extracolsep{\fill}}lccc@{}}
        \toprule
        Method & \makecell{Table Tennis\\Success (\%)$\uparrow$}
        & \makecell{Conveyor\\Success (\%)$\uparrow$}
        & \makecell{Obs. Age (ms)$\downarrow$\\P50 / P95} \\
        \midrule
        Sync & 0 & 0 & 308.75 / 456.24 \\
        Naive Async & 5 & 20 & 235.59 / 307.34 \\
        RTC & 0 & 0 & 465.30 / 640.84 \\
        VLASH & 25 & 90 & 244.12 / 331.24 \\
        \method{} & \textbf{75} & \textbf{95} & \textbf{118.06 / 166.95} \\
        \bottomrule
    \end{tabular*}
\end{table}

\begin{table}[t]
    \centering
    \caption{\textbf{Real-world tennis-ball grasping.}
    Success rates are computed over 20 trials per method.
    $E_{\mathrm{HF}}$ and TCP jerk RMS characterize trajectory oscillations
    and motion smoothness, respectively. Bold indicates the best result
    for each metric.}
    \label{tab:real-grasping}
    \begin{tabular*}{\linewidth}{@{\extracolsep{\fill}}lccc@{}}
        \toprule
        Method & \makecell{Success\\(\%)$\uparrow$}
        & \makecell{$E_{\mathrm{HF}}\downarrow$\\(mm$^2$)}
        & \makecell{TCP jerk$\downarrow$\\(mm/s$^3$)} \\
        \midrule
        Sync & 40 & 14.45 & 1983 \\
        Naive Async & \textbf{45} & 19.43 & 3208 \\
        RTC & 40 & 101.70 & 8398 \\
        \method{} & \textbf{45} & \textbf{7.01} & \textbf{1800} \\
        \bottomrule
    \end{tabular*}
\end{table}

\paragraph{Results.}
As shown in Table~\ref{tab:real-world}, \method{} achieves the highest
success rates on both real-world tasks, reaching 75\% on table-tennis return
and 95\% on conveyor-belt grasping. Its success-rate improvement is
particularly pronounced in table-tennis return, where it outperforms the
strongest baseline by 50 percentage points. \method{} also achieves the
lowest observation age among all evaluated methods, approximately 50\%
and 46\% lower than Naive Async at P50 and P95, respectively. This
reduction in observation age is particularly beneficial for table-tennis
return, where strict timing requirements make task success highly
sensitive to inference latency. Sync relies on increasingly stale
observations as it executes each action chunk, while RTC incurs
additional update latency through vector--Jacobian products in its
inference-time guidance. In contrast, \method{} runs VLM context
encoding and rolling action updates in separate workers, allowing
action inference to continue without waiting for each VLM update and
maintaining consistently fresher observations.

On tennis-ball grasping (Table~\ref{tab:real-grasping}), \method{} achieves comparable task success while
exhibiting the lowest trajectory variation across both $E_{\mathrm{HF}}$
and TCP jerk RMS. Together, these metrics indicate fewer high-frequency
oscillations and fewer abrupt motion changes during execution. The
chunk-based baselines update actions less frequently, so latency can produce
larger discontinuities when switching to a newly generated chunk. \method{}
instead streams actions through rolling inference and continuously refines
future actions across control cycles, resulting in smoother transitions
and more stable execution. We also find that successful grasps occur at similar ball locations
across methods. This is likely because the methods share similar
manipulation capabilities, while \method{} mainly improves how the
actions are executed.

\subsection{Analysis and Ablation Studies}
\label{sec:ablations}

\begin{figure}[!t]
    \centering
    \includegraphics[width=\linewidth]{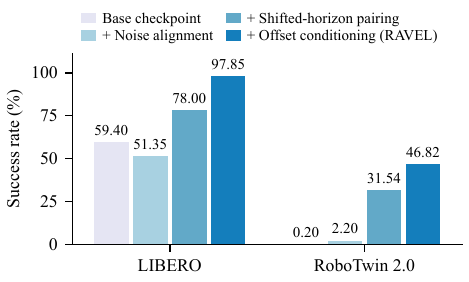}
    \caption{Training-time alignment for $\pi_{0.5}$ under rolling inference.
    All variants use 2,000 episodes on LIBERO and 5,000 episodes on RoboTwin~2.0.}
    \label{fig:training-alignment}
\end{figure}

\paragraph{Ablation of Training-Time Alignment Components.}
We ablate the training components of \method{} using $\pi_{0.5}$ on LIBERO
and RoboTwin~2.0, with the rolling inference procedure fixed across all four
variants. Figure~\ref{fig:training-alignment} shows that directly applying rolling
inference to the pretrained checkpoint performs poorly, and matching the
training noise profile to the rolling buffer is insufficient to recover
performance. Training on action horizons shifted relative to the observation
improves success rates to 78.00\% on LIBERO and 31.54\% on RoboTwin~2.0.
However, without temporal-offset conditioning, the model is not told which
future horizon to predict from a given observation. Providing this offset
further improves success rates to 97.85\% and 46.82\%, respectively.
These results suggest that adapting a pretrained policy to rolling inference
requires both training on shifted action horizons and conditioning on their
temporal offsets.

\begin{figure}[!t]
    \centering
    \includegraphics[width=\linewidth]{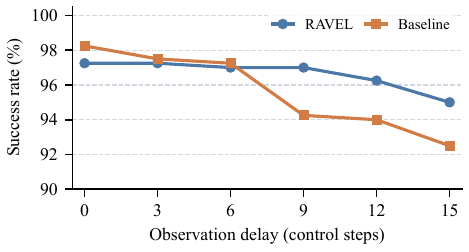}
    \caption{Robustness to delayed observations on LIBERO with X-VLA under
    synchronous rollouts. Each point uses 400 trials.}
    \label{fig:libero-observation-delay}
\end{figure}

\paragraph{Temporal Alignment under Inference Latency.}
Inference latency introduces a delay between observation capture and action
execution. Temporal-offset conditioning informs the action expert of this delay,
allowing it to predict actions for the intended execution time. We examine this
temporal alignment using controlled observation delays on LIBERO and an offset
ablation under asynchronous execution on DOM.

We first simulate inference latency in a synchronous environment using delayed
observations. On LIBERO with X-VLA, we vary the observation delay from 0 to
15 control steps. Across these delays, \method{} maintains 95.00--97.25\% success,
whereas the baseline decreases from 98.25\% to 92.50\%
(Figure~\ref{fig:libero-observation-delay}). This indicates that temporal-offset conditioning
improves robustness to the temporal misalignment caused by inference latency.

We further evaluate the offset under asynchronous execution on DOM, where
continuously changing object states make the policy more sensitive to temporal
misalignment. Because VLM encoding takes multiple control steps, a newly available
context is already stale. \method{} uses the actual offset $p=t_a-t_o$, while
the comparison variant resets the offset to zero whenever a new context becomes
available, thereby ignoring the VLM computation delay. With the checkpoint and
non-blocking runtime fixed, success drops from 44.89\% to 16.67\% over 450 paired
episodes. This result highlights the importance of including VLM computation
time in the temporal offset to align action predictions with their intended
execution times.

\begin{figure}[!t]
    \centering
    \includegraphics[width=0.93\linewidth]{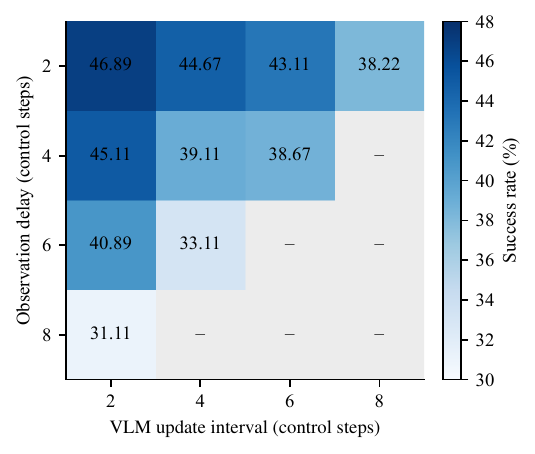}
    \caption{DOM success rates with controlled observation delays and VLM
    update intervals, both measured in control steps.
    Each setting uses the same 450 episodes.}
    \label{fig:dom-observation-delay}
\end{figure}

\paragraph{Effect of Observation Delay and VLM Update Interval.}
We investigate how the freshness and update frequency of VLM context affect dynamic task performance.
We use a fixed DynamicVLA checkpoint on DOM and pause the simulator during inference,
allowing us to vary these two factors independently of actual inference time.
Observation delay specifies how many frames earlier the image used for VLM inference was captured.
The VLM update interval is the number of control steps between consecutive VLM forward passes that refresh the context.
Between updates, the action expert continues to use the existing context with the corresponding observation-to-target offset.
We vary both factors over 2, 4, 6, and 8 control steps and evaluate each setting on the same 450 episodes.

With the update interval fixed at two steps, increasing observation delay from two to eight steps
reduces success from 46.89\% to 31.11\% (Figure~\ref{fig:dom-observation-delay}).
With observation delay fixed at two steps, increasing the update interval from two to eight steps
reduces success to 38.22\%.
These results demonstrate that both observation delay and the VLM update interval affect dynamic task performance:
fresher observations and more frequent context updates lead to higher success rates.

\section{Limitations and Future Work}
\label{sec:conclusion}

\method{} provides a rolling asynchronous inference framework for improving the
responsiveness of flow-based VLA models. It currently requires streaming-aligned
training, while temporal-offset conditioning and FOP involve architecture-specific
modifications to the action expert because existing VLA models do not share a
unified action-expert design. The framework is presently formulated for flow-based
VLAs, and its action-generation latency is bounded below by the time required for
a single action-expert forward pass. FOP improves observation freshness by providing the action expert
with current observations, but the subsequent delay between action prediction
and physical execution remains in real-world systems, potentially reducing its
benefits. Future work could explore more general and efficient streaming inference
across VLA architectures, improving compatibility with diverse action-expert
designs or lowering the computational cost of action generation. Future work
could also investigate tighter integration between streaming action generation,
robot motion planning, and execution feedback, accounting for physical delays
and the robot's evolving state for more responsive closed-loop control.

\medskip
\noindent We used ChatGPT and OpenAI Codex to assist with implementation,
literature searches, manuscript preparation, figure preparation, and analysis
of experimental results.

{\small\bibliographystyle{ieeenat_fullname}
\bibliography{references}}

\begin{thebibliography}{42}
\providecommand{\natexlab}[1]{#1}
\providecommand{\url}[1]{\texttt{#1}}
\expandafter\ifx\csname urlstyle\endcsname\relax
  \providecommand{\doi}[1]{doi: #1}\else
  \providecommand{\doi}{doi: \begingroup \urlstyle{rm}\Url}\fi

\bibitem[Bi et~al.(2025)Bi, Tan, Xie, Wang, Huang, Liu, Zhao, Feng, Xiang,
  Rong, Zhao, Liu, Su, Ma, Su, and Zhu]{motusPi05Robotwin2}
Hongzhe Bi, Hengkai Tan, Shenghao Xie, Zeyuan Wang, Shuhe Huang, Haitian Liu,
  Ruowen Zhao, Yao Feng, Chendong Xiang, Yinze Rong, Hongyan Zhao, Hanyu Liu,
  Zhizhong Su, Lei Ma, Hang Su, and Jun Zhu.
\newblock Motus: A unified latent action world model, 2025.

\bibitem[Bjorck et~al.(2025)Bjorck, Casta{\~n}eda, Cherniadev, Da, Ding, Fan,
  Fang, Fox, Hu, Huang, et~al.]{bjorck2025grootn1}
Johan Bjorck, Fernando Casta{\~n}eda, Nikita Cherniadev, Xingye Da, Runyu Ding,
  Linxi Fan, Yu Fang, Dieter Fox, Fengyuan Hu, Spencer Huang, et~al.
\newblock Gr00t n1: An open foundation model for generalist humanoid robots.
\newblock \emph{arXiv preprint arXiv:2503.14734}, 2025.

\bibitem[Black et~al.(2024)Black, Brown, Driess, Esmail, Equi, Finn, Fusai,
  Groom, Hausman, Ichter, et~al.]{black2024pi0}
Kevin Black, Noah Brown, Danny Driess, Adnan Esmail, Michael Equi, Chelsea
  Finn, Niccolo Fusai, Lachy Groom, Karol Hausman, Brian Ichter, et~al.
\newblock {$\pi_0$}: A vision-language-action flow model for general robot
  control.
\newblock \emph{arXiv preprint arXiv:2410.24164}, 2024.

\bibitem[Black et~al.(2025{\natexlab{a}})Black, Galliker, and
  Levine]{black2025rtc}
Kevin Black, Manuel Galliker, and Sergey Levine.
\newblock Real-time execution of action chunking flow policies.
\newblock In \emph{Advances in Neural Information Processing Systems}, pages
  33383--33407. Curran Associates, Inc., 2025{\natexlab{a}}.

\bibitem[Black et~al.(2025{\natexlab{b}})Black, Ren, Equi, and
  Levine]{black2025trainingrtc}
Kevin Black, Allen~Z Ren, Michael Equi, and Sergey Levine.
\newblock Training-time action conditioning for efficient real-time chunking.
\newblock \emph{arXiv preprint arXiv:2512.05964}, 2025{\natexlab{b}}.

\bibitem[Chen et~al.(2024)Chen, Mart{\'\i}~Mons{\'o}, Du, Simchowitz, Tedrake,
  and Sitzmann]{chen2024diffusionforcing}
Boyuan Chen, Diego Mart{\'\i}~Mons{\'o}, Yilun Du, Max Simchowitz, Russ
  Tedrake, and Vincent Sitzmann.
\newblock Diffusion forcing: Next-token prediction meets full-sequence
  diffusion.
\newblock \emph{Advances in Neural Information Processing Systems},
  37:\penalty0 24081--24125, 2024.

\bibitem[Chen et~al.(2025{\natexlab{a}})Chen, Liu, Ma, Ma, Ma, Wu, Chen, Zhong,
  Wang, Li, et~al.]{chen2025falcon}
Haojun Chen, Minghao Liu, Chengdong Ma, Xiaojian Ma, Zailin Ma, Huimin Wu,
  Yuanpei Chen, Yifan Zhong, Mingzhi Wang, Qing Li, et~al.
\newblock Falcon: Fast visuomotor policies via partial denoising.
\newblock \emph{arXiv preprint arXiv:2503.00339}, 2025{\natexlab{a}}.

\bibitem[Chen et~al.(2025{\natexlab{b}})Chen, Chen, Chen, Cai, Liu, Li, Liang,
  Lin, Ge, Gu, et~al.]{chen2025robotwin2}
Tianxing Chen, Zanxin Chen, Baijun Chen, Zijian Cai, Yibin Liu, Zixuan Li,
  Qiwei Liang, Xianliang Lin, Yiheng Ge, Zhenyu Gu, et~al.
\newblock Robotwin 2.0: A scalable data generator and benchmark with strong
  domain randomization for robust bimanual robotic manipulation.
\newblock \emph{arXiv preprint arXiv:2506.18088}, 2025{\natexlab{b}}.

\bibitem[Chen et~al.(2025{\natexlab{c}})Chen, Yuan, Mu, and Su]{chen2025rnrdp}
Zhuoqun Chen, Xiu Yuan, Tongzhou Mu, and Hao Su.
\newblock Responsive noise-relaying diffusion policy: Responsive and efficient
  visuomotor control.
\newblock \emph{arXiv preprint arXiv:2502.12724}, 2025{\natexlab{c}}.

\bibitem[Chi et~al.(2025)Chi, Xu, Feng, Cousineau, Du, Burchfiel, Tedrake, and
  Song]{chi2023diffusionpolicy}
Cheng Chi, Zhenjia Xu, Siyuan Feng, Eric Cousineau, Yilun Du, Benjamin
  Burchfiel, Russ Tedrake, and Shuran Song.
\newblock Diffusion policy: Visuomotor policy learning via action diffusion.
\newblock \emph{The International Journal of Robotics Research}, 44\penalty0
  (10-11):\penalty0 1684--1704, 2025.

\bibitem[Du et~al.(2026)Du, Yan, Wu, Xu, Zhang, Wang, Guo, Qian, He, Wang,
  et~al.]{du2026cfvla}
Fan Du, Feng Yan, Jianxiong Wu, Xinrun Xu, Weiye Zhang, Weinong Wang, Yu Guo,
  Bin Qian, Zhihai He, Fei Wang, et~al.
\newblock Cf-vla: Efficient coarse-to-fine action generation for
  vision-language-action policies.
\newblock \emph{arXiv preprint arXiv:2604.24622}, 2026.

\bibitem[Guo and Liu(2026)]{guo2026reflex}
Yuanchun Guo and Bingyan Liu.
\newblock Reflex: Real-time vla control through streaming inference.
\newblock \emph{arXiv preprint arXiv:2607.14695}, 2026.

\bibitem[H{\o}eg et~al.(2024)H{\o}eg, Du, and Egeland]{hoeg2024sdp}
Sigmund~H H{\o}eg, Yilun Du, and Olav Egeland.
\newblock Streaming diffusion policy: Fast policy synthesis with variable noise
  diffusion models.
\newblock \emph{arXiv preprint arXiv:2406.04806}, 2024.

\bibitem[Intelligence et~al.(2025)Intelligence, Black, Brown, Darpinian,
  Dhabalia, Driess, Esmail, Equi, Finn, Fusai, et~al.]{black2025pi05}
Physical Intelligence, Kevin Black, Noah Brown, James Darpinian, Karan
  Dhabalia, Danny Driess, Adnan Esmail, Michael Equi, Chelsea Finn, Niccolo
  Fusai, et~al.
\newblock {$\pi_{0.5}$}: a vision-language-action model with open-world
  generalization.
\newblock \emph{arXiv preprint arXiv:2504.16054}, 2025.

\bibitem[Jiang et~al.(2025)Jiang, Fang, Roy, Lozano-P{\'e}rez, Kaelbling, and
  Ancha]{jiang2025streamingflow}
Sunshine Jiang, Xiaolin Fang, Nicholas Roy, Tom{\'a}s Lozano-P{\'e}rez,
  Leslie~Pack Kaelbling, and Siddharth Ancha.
\newblock Streaming flow policy: Simplifying diffusion/flow-matching policies
  by treating action trajectories as flow trajectories.
\newblock \emph{arXiv preprint arXiv:2505.21851}, 2025.

\bibitem[Jung et~al.(2025)Jung, Ahn, Kim, Jang, Kim, Yoo, and Ko]{jung2025rdp}
Chanhyuk Jung, Dasom Ahn, Sangwon Kim, In-su Jang, Kwang-Ju Kim, Sungkeun Yoo,
  and Byoung~Chul Ko.
\newblock Rolling diffusion policy for robotic action prediction: Enhancing
  efficiency and temporal awareness.
\newblock In \emph{ICRA 2025 Workshop on Foundation Models and Neuro-Symbolic
  AI for Robotics}, 2025.

\bibitem[Kim et~al.(2024)Kim, Pertsch, Karamcheti, Xiao, Balakrishna, Nair,
  Rafailov, Foster, Lam, Sanketi, et~al.]{kim2024openvla}
Moo~Jin Kim, Karl Pertsch, Siddharth Karamcheti, Ted Xiao, Ashwin Balakrishna,
  Suraj Nair, Rafael Rafailov, Ethan Foster, Grace Lam, Pannag Sanketi, et~al.
\newblock Openvla: An open-source vision-language-action model.
\newblock \emph{arXiv preprint arXiv:2406.09246}, 2024.

\bibitem[Li et~al.(2024)Li, Liang, Wang, Luo, Chen, Liao, Wei, Deng, Xu, Zhang,
  et~al.]{li2024cogact}
Qixiu Li, Yaobo Liang, Zeyu Wang, Lin Luo, Xi Chen, Mozheng Liao, Fangyun Wei,
  Yu Deng, Sicheng Xu, Yizhong Zhang, et~al.
\newblock Cogact: A foundational vision-language-action model for synergizing
  cognition and action in robotic manipulation.
\newblock \emph{arXiv preprint arXiv:2411.19650}, 2024.

\bibitem[Lipman et~al.(2022)Lipman, Chen, Ben-Hamu, Nickel, and
  Le]{lipman2023flow}
Yaron Lipman, Ricky~TQ Chen, Heli Ben-Hamu, Maximilian Nickel, and Matt Le.
\newblock Flow matching for generative modeling.
\newblock \emph{arXiv preprint arXiv:2210.02747}, 2022.

\bibitem[Liu et~al.(2023)Liu, Zhu, Gao, Feng, Liu, Zhu, and
  Stone]{liu2023libero}
Bo Liu, Yifeng Zhu, Chongkai Gao, Yihao Feng, Qiang Liu, Yuke Zhu, and Peter
  Stone.
\newblock Libero: Benchmarking knowledge transfer for lifelong robot learning.
\newblock \emph{Advances in Neural Information Processing Systems},
  36:\penalty0 44776--44791, 2023.

\bibitem[Liu et~al.(2025)Liu, Wu, Li, Tan, Chen, Wang, Xu, Su, and
  Zhu]{liu2024rdt}
Songming Liu, Lingxuan Wu, Bangguo Li, Hengkai Tan, Huayu Chen, Zhengyi Wang,
  Ke Xu, Hang Su, and Jun Zhu.
\newblock Rdt-1b: a diffusion foundation model for bimanual manipulation.
\newblock In \emph{International Conference on Learning Representations}, pages
  29982--30009, 2025.

\bibitem[Liu et~al.(2026)Liu, Yu, Zhao, Li, Zhang, Li, Wu, Hu, Xie, Guo,
  et~al.]{liu2026legato}
Yufeng Liu, Hang Yu, Juntu Zhao, Bocheng Li, Di Zhang, Mingzhu Li, Wenxuan Wu,
  Yingdong Hu, Junyuan Xie, Junliang Guo, et~al.
\newblock Learning native continuation for action chunking flow policies.
\newblock \emph{arXiv preprint arXiv:2602.12978}, 2026.

\bibitem[Liu et~al.(2022)Liu, Mao, Wu, Feichtenhofer, Darrell, and
  Xie]{liu2022convnet}
Zhuang Liu, Hanzi Mao, Chao-Yuan Wu, Christoph Feichtenhofer, Trevor Darrell,
  and Saining Xie.
\newblock A convnet for the 2020s.
\newblock In \emph{2022 IEEE/CVF conference on computer vision and pattern
  recognition (CVPR)}, pages 11966--11976. IEEE, 2022.

\bibitem[Lu et~al.(2026)Lu, Liu, Fan, Yang, Hou, Li, Ding, and
  Zhao]{lu2026faster}
Yuxiang Lu, Zhe Liu, Xianzhe Fan, Zhenya Yang, Jinghua Hou, Junyi Li, Kaixin
  Ding, and Hengshuang Zhao.
\newblock Faster: Rethinking real-time flow vlas.
\newblock \emph{arXiv preprint arXiv:2603.19199}, 2026.

\bibitem[Ma et~al.(2025)Ma, Zhou, Yang, Wang, and Fan]{ma2025running}
Yunchao Ma, Yizhuang Zhou, Yunhuan Yang, Tiancai Wang, and Haoqiang Fan.
\newblock Running vlas at real-time speed.
\newblock \emph{arXiv preprint arXiv:2510.26742}, 2025.

\bibitem[Ruhe et~al.(2024)Ruhe, Heek, Salimans, and Hoogeboom]{ruhe2024rolling}
David Ruhe, Jonathan Heek, Tim Salimans, and Emiel Hoogeboom.
\newblock Rolling diffusion models.
\newblock \emph{arXiv preprint arXiv:2402.09470}, 2024.

\bibitem[Sendai et~al.(2025)Sendai, Alvarez, Matsushima, Matsuo, and
  Iwasawa]{sendai2025a2c2}
Kohei Sendai, Maxime Alvarez, Tatsuya Matsushima, Yutaka Matsuo, and Yusuke
  Iwasawa.
\newblock Leave no observation behind: Real-time correction for vla action
  chunks.
\newblock \emph{arXiv preprint arXiv:2509.23224}, 2025.

\bibitem[Shao et~al.(2025)Shao, Li, Zhang, Zhang, Liu, Chen, and
  Nie]{shao2025vlasurvey}
Rui Shao, Wei Li, Lingsen Zhang, Renshan Zhang, Zhiyang Liu, Ran Chen, and
  Liqiang Nie.
\newblock Large vlm-based vision-language-action models for robotic
  manipulation: A survey.
\newblock \emph{arXiv preprint arXiv:2508.13073}, 2025.

\bibitem[Shukor et~al.(2025)Shukor, Aubakirova, Capuano, Kooijmans, Palma,
  Zouitine, Aractingi, Pascal, Russi, Marafioti, Alibert, Cord, Wolf, and
  Cadene]{shukor2025smolvla}
Mustafa Shukor, Dana Aubakirova, Francesco Capuano, Pepijn Kooijmans, Steven
  Palma, Adil Zouitine, Michel Aractingi, Caroline Pascal, Martino Russi,
  Andres Marafioti, Simon Alibert, Matthieu Cord, Thomas Wolf, and Remi Cadene.
\newblock Smolvla: A vision-language-action model for affordable and efficient
  robotics, 2025.

\bibitem[Sojib et~al.(2026)Sojib, Arthanat, and
  Begum]{sojib2026efficientmetric}
Noushad Sojib, Sajay Arthanat, and Momotaz Begum.
\newblock An efficient metric for data quality measurement in imitation
  learning.
\newblock \emph{arXiv preprint arXiv:2605.01544}, 2026.

\bibitem[Tang et~al.(2025)Tang, Sun, Zhao, Yang, Lin, Zhang, Hou, Lu, Liu, and
  Han]{tang2025vlash}
Jiaming Tang, Yufei Sun, Yilong Zhao, Shang Yang, Yujun Lin, Zhuoyang Zhang,
  James Hou, Yao Lu, Zhijian Liu, and Song Han.
\newblock Vlash: Real-time vlas via future-state-aware asynchronous inference.
\newblock \emph{arXiv preprint arXiv:2512.01031}, 2025.

\bibitem[Team et~al.(2025)Team, Abeyruwan, Ainslie, Alayrac, Arenas, Armstrong,
  Balakrishna, Baruch, Bauza, Blokzijl, et~al.]{team2025gemini}
Gemini~Robotics Team, Saminda Abeyruwan, Joshua Ainslie, Jean-Baptiste Alayrac,
  Montserrat~Gonzalez Arenas, Travis Armstrong, Ashwin Balakrishna, Robert
  Baruch, Maria Bauza, Michiel Blokzijl, et~al.
\newblock Gemini robotics: Bringing ai into the physical world.
\newblock \emph{arXiv preprint arXiv:2503.20020}, 2025.

\bibitem[Wang et~al.(2026{\natexlab{a}})Wang, Zhang, Yan, Shang, Kompella, and
  Liu]{wang2026remac}
Haoxuan Wang, Gengyu Zhang, Yan Yan, Yuzhang Shang, Ramana Kompella, and Gaowen
  Liu.
\newblock Real-time robot execution with masked action chunking.
\newblock In \emph{International Conference on Learning Representations}, pages
  120161--120179, 2026{\natexlab{a}}.

\bibitem[Wang et~al.(2026{\natexlab{b}})Wang, Li, Guan, Ye, Xie, Liu, Chen,
  Liang, Zhang, Hu, et~al.]{wang2026qwenvla}
Qiuyue Wang, Mingsheng Li, Jian Guan, Jinhui Ye, Sicheng Xie, Yitao Liu, Junhao
  Chen, Zhixuan Liang, Jie Zhang, Xintong Hu, et~al.
\newblock Qwen-vla: Unifying vision-language-action modeling across tasks,
  environments, and robot embodiments.
\newblock \emph{arXiv preprint arXiv:2605.30280}, 2026{\natexlab{b}}.

\bibitem[Wu et~al.(2024)Wu, Shentu, Yi, Lin, and Abbeel]{wu2023gello}
Philipp Wu, Yide Shentu, Zhongke Yi, Xingyu Lin, and Pieter Abbeel.
\newblock Gello: A general, low-cost, and intuitive teleoperation framework for
  robot manipulators, 2024.

\bibitem[Wu et~al.(2026)Wu, Lu, Wang, Yang, Liu, Wang, Zhu, Sun, Wang, Ma,
  et~al.]{wu2026lingbotvla}
Wei Wu, Fan Lu, Yunnan Wang, Shuai Yang, Shi Liu, Fangjing Wang, Qian Zhu, He
  Sun, Yong Wang, Shuailei Ma, et~al.
\newblock A pragmatic vla foundation model.
\newblock \emph{arXiv preprint arXiv:2601.18692}, 2026.

\bibitem[Xie et~al.(2026)Xie, Wen, Zheng, Chen, Hong, Diao, and
  Liu]{xie2026dynamicvla}
Haozhe Xie, Beichen Wen, Jiarui Zheng, Zhaoxi Chen, Fangzhou Hong, Haiwen Diao,
  and Ziwei Liu.
\newblock Dynamicvla: A vision-language-action model for dynamic object
  manipulation.
\newblock \emph{arXiv preprint arXiv:2601.22153}, 2026.

\bibitem[Zhang et~al.(2024)Zhang, Guo, Chen, Wang, Hu, Shi, and
  Chen]{zhang2025hirt}
Jianke Zhang, Yanjiang Guo, Xiaoyu Chen, Yen-Jen Wang, Yucheng Hu, Chengming
  Shi, and Jianyu Chen.
\newblock Hirt: Enhancing robotic control with hierarchical robot transformers.
\newblock \emph{arXiv preprint arXiv:2410.05273}, 2024.

\bibitem[Zhao et~al.(2023)Zhao, Kumar, Levine, and Finn]{zhao2023act}
Tony~Z Zhao, Vikash Kumar, Sergey Levine, and Chelsea Finn.
\newblock Learning fine-grained bimanual manipulation with low-cost hardware.
\newblock \emph{arXiv preprint arXiv:2304.13705}, 2023.

\bibitem[Zhao et~al.(2025)Zhao, Zhao, Cheng, Yao, Wen, and
  Gao]{zhao2025vlarail}
Yongsheng Zhao, Lei Zhao, Baoping Cheng, Gongxin Yao, Xuanzhang Wen, and Han
  Gao.
\newblock Vla-rail: A real-time asynchronous inference linker for vla models
  and robots.
\newblock \emph{arXiv preprint arXiv:2512.24673}, 2025.

\bibitem[Zheng et~al.(2026)Zheng, Li, Wang, Liu, Kang, Feng, Zheng, Zou, Chen,
  Zeng, et~al.]{zheng2025xvla}
Jinliang Zheng, Jianxiong Li, Zhihao Wang, Dongxiu Liu, Xirui Kang, Yuchun
  Feng, Yinan Zheng, Jiayin Zou, Yilun Chen, Jia Zeng, et~al.
\newblock X-vla: Soft-prompted transformer as scalable cross-embodiment
  vision-language-action model.
\newblock In \emph{International Conference on Learning Representations}, pages
  60580--60606, 2026.

\bibitem[Zitkovich et~al.(2023)Zitkovich, Yu, Xu, Xu, Xiao, Xia, Wu, Wohlhart,
  Welker, Wahid, Vuong, Vanhoucke, Tran, Soricut, Singh, Singh, Sermanet,
  Sanketi, Salazar, Ryoo, Reymann, Rao, Pertsch, Mordatch, Michalewski, Lu,
  Levine, Lee, Lee, Leal, Kuang, Kalashnikov, Julian, Joshi, Irpan, Ichter,
  Hsu, Herzog, Hausman, Gopalakrishnan, Fu, Florence, Finn, Dubey, Driess,
  Ding, Choromanski, Chen, Chebotar, Carbajal, Brown, Brohan, Arenas, and
  Han]{brohan2023rt2}
Brianna Zitkovich, Tianhe Yu, Sichun Xu, Peng Xu, Ted Xiao, Fei Xia, Jialin Wu,
  Paul Wohlhart, Stefan Welker, Ayzaan Wahid, Quan Vuong, Vincent Vanhoucke,
  Huong Tran, Radu Soricut, Anikait Singh, Jaspiar Singh, Pierre Sermanet,
  Pannag~R. Sanketi, Grecia Salazar, Michael~S. Ryoo, Krista Reymann, Kanishka
  Rao, Karl Pertsch, Igor Mordatch, Henryk Michalewski, Yao Lu, Sergey Levine,
  Lisa Lee, Tsang-Wei~Edward Lee, Isabel Leal, Yuheng Kuang, Dmitry
  Kalashnikov, Ryan Julian, Nikhil~J. Joshi, Alex Irpan, Brian Ichter, Jasmine
  Hsu, Alexander Herzog, Karol Hausman, Keerthana Gopalakrishnan, Chuyuan Fu,
  Pete Florence, Chelsea Finn, Kumar~Avinava Dubey, Danny Driess, Tianli Ding,
  Krzysztof~Marcin Choromanski, Xi Chen, Yevgen Chebotar, Justice Carbajal,
  Noah Brown, Anthony Brohan, Montserrat~Gonzalez Arenas, and Kehang Han.
\newblock {RT-2}: Vision-language-action models transfer web knowledge to
  robotic control.
\newblock In \emph{Proceedings of The 7th Conference on Robot Learning}, pages
  2165--2183, 2023.

\end{thebibliography}
\clearpage
\appendix
\section{Simulation Implementation Details}

\subsection{LIBERO and RoboTwin~2.0}
\label{app:sync-details}

We use the official X-VLA checkpoints~\citep{zheng2025xvla} for LIBERO and
RoboTwin~2.0, and the official OpenPI \texttt{pi05\_libero} checkpoint for
$\pi_{0.5}$ on LIBERO~\citep{black2025pi05}. For $\pi_{0.5}$ on RoboTwin~2.0, we use the checkpoint released by
Motus Robotics~\citep{motusPi05Robotwin2}.
For $\pi_{0.5}$ on LIBERO, we convert
OpenPI's released JAX \texttt{pi05\_libero} weights to PyTorch using its
checkpoint-conversion script. All four settings use PyTorch inference.

All four configurations use $K=10$ buffer slots. Each slot contains $h$
consecutive low-level actions, giving a total horizon of $H=Kh$;
a rolling update releases one slot.
We do not perform any additional initialization. The first target action
is predicted directly with a single denoising update. As rolling updates proceed, subsequent actions undergo
progressively more denoising steps before execution, until each action
receives $K$ updates in steady state.

The temporal offset $\delta_p=p/K$ encodes the age of the VLM context.
For $\pi_{0.5}$, we encode $\delta_p$ with an MLP and add the resulting
embedding to the flow-timestep embedding, jointly conditioning the action
expert on context age and denoising progress. For X-VLA,
we concatenate $\delta_p$ with the noisy action, proprioception, and
flow-timestep embedding, and feed the combined features to the action encoder.
All action slots share the same offset and use their individual flow timesteps.

We fine-tune the released checkpoints described above to obtain the four
\method{} models. For $\pi_{0.5}$, we freeze the vision--language
backbone and train the action expert, action input/output projections, flow-timestep MLP, and
added temporal-offset MLP. For X-VLA, we freeze the vision--language
backbone and the soft prompts, and train the action Transformer and action input/output heads,
including the input mapping extended to accept the temporal offset.
Table~\ref{tab:sync-training-config} summarizes the training and evaluation
configurations for all four settings.

\begin{table*}[!ht]
    \caption{Training and evaluation configurations for synchronous evaluation. Batch size is
    global; steps denote fine-tuning steps for \method{}. EE denotes
    end-effector pose.}
    \label{tab:sync-training-config}
    \begin{center}
    \begin{tabular*}{\linewidth}{@{\extracolsep{\fill}}lllrrrr@{}}
        \toprule
        Benchmark & Backbone & Action representation & $H$ & $h$ & Steps & Batch \\
        \midrule
        LIBERO & $\pi_{0.5}$ & Relative EE pose & 10 & 1 & 5K & 128 \\
        LIBERO & X-VLA & Absolute EE pose & 30 & 3 & 5K & 128 \\
        RoboTwin~2.0 & $\pi_{0.5}$ & Relative joint position & 30 & 3 & 10K & 128 \\
        RoboTwin~2.0 & X-VLA & Absolute EE pose & 30 & 3 & 10K & 128 \\
        \bottomrule
    \end{tabular*}
    \end{center}
\end{table*}

During synchronous evaluation, each policy query uses
one VLM prefill and ten action-expert forward passes. The baseline performs
ten full-horizon denoising steps, while \method{} performs ten rolling updates
to produce $H$ actions.

For $\pi_{0.5}$ on RoboTwin~2.0, the model predicts relative joint actions, expressed as offsets from
the current robot state. Rolling denoising cannot directly reuse these
relative actions across observations because their reference state changes as the
robot moves. Interpreting a stored relative action with respect to a new state would
change its intended joint target. We therefore store normalized absolute
joint targets in the clean buffer and convert them to relative actions with respect to
the latest state before each action-expert update. The predicted clean
relative actions are converted back to absolute targets before being written to the
buffer. The paired noise remains in model coordinates. This conversion
preserves the intended targets across rolling updates.

\subsection{Dynamic Object Manipulation}
\label{app:dom-details}

\paragraph{DynamicVLA inference latency.}
Table~\ref{tab:dynamicvla-latency} reports local RTX~4090 inference latency.
With the same model configuration as the official release, our median full-model
inference latency is 370.7\,ms, compared with the 226\,ms reported on RTX~A6000
by the original paper~\citep{xie2026dynamicvla}.
Because DOM involves dynamic manipulation, slower inference in our local
setup likely contributes to the gap between our reproduced success rates
and those reported in the original paper.

\begin{table}[t]
    \centering
    \caption{DynamicVLA inference latency on an NVIDIA RTX~4090.
    P50 and P95 denote the median and 95th percentile.}
    \label{tab:dynamicvla-latency}
    \begin{tabular*}{\linewidth}{@{\extracolsep{\fill}}lrr@{}}
        \toprule
        Component / inference path & P50 (ms) & P95 (ms) \\
        \midrule
        VLM encoding & 63.47 & 84.34 \\
        Action expert & 31.75 & 48.75 \\
        Action expert with FOP & 41.4 & 58.6 \\
        Full model & 370.7 & 433.5 \\
        \bottomrule
    \end{tabular*}
\end{table}

\paragraph{\method{}.}
We extend DynamicVLA's action expert with temporal-offset conditioning.
The offset is concatenated with the action and flow-timestep embeddings
before the action-time MLP, and its sinusoidal embedding conditions a
gated residual adapter on the resulting action tokens.
FOP comprises a ConvNeXt-Base image encoder~\citep{liu2022convnet},
visual/state projections, camera/state embeddings, and gated cross-attention
adapters that inject current observations into the action expert.

We freeze the VLM and train the action expert and temporal adapter for
100K steps with global batch size 15, learning rates $10^{-6}$ and
$3\times10^{-5}$, respectively, and progressive noise. FOP training starts from this checkpoint. With the policy and image encoder
frozen, we train its projections, embeddings, and adapters for 50K steps
with global batch size 120 and learning rate $10^{-4}$.

DynamicVLA predicts relative actions: end-effector pose targets expressed
relative to the reference robot state, with absolute gripper commands. To reuse the buffer
across changing reference states, we apply the same coordinate conversion
as for $\pi_{0.5}$ on RoboTwin~2.0 (Appendix~\ref{app:sync-details}).

\paragraph{DOM execution timeline.}
DOM advances the environment at 25\,Hz during policy inference and
compensates for simulation slowdown when determining result availability.
We follow this timing convention in our evaluation.

We emulate asynchronous context availability within a single policy
process by executing VLM encoding and action-expert updates serially.
Each inference cycle first updates the rolling action buffer using the
previous VLM context, then encodes the query observation, and finally
updates the buffer using the newly computed context.
The temporal offset is computed from each action block's target step and
the source observation used to generate its VLM context.
Serial execution allows us to apply the environment's time compensation
at each stage while preserving this order.

The control period is 40\,ms, each action block contains two actions
(80\,ms), and the nominal VLM query interval is four control steps (160\,ms).
A query expires when the simulator reaches four control steps after its
source observation. On expiry, the runtime stops further stages and
suppresses action blocks that have not yet been released; blocks already
released remain scheduled. If no valid action is available for a control
step, the simulator holds the previous command. Exceeding the cycle budget
can therefore reduce the frequency of fresh action updates.

Table~\ref{tab:dom-serial-cycle} reports the P50 and P95 computation times
of the full serial cycle, with and without FOP. Overrun denotes the
percentage of complete computation cycles exceeding the 160\,ms budget.
Most complete computation cycles fit within this budget, with overrun rates
of 2.52\% without FOP and 11.39\% with FOP. When a query expires, pending
action updates are discarded, and the robot holds its previous command if
no scheduled action is available. This reduces the frequency of fresh
corrections and can impair tracking of moving objects. With FOP, serial cycles take longer to compute and exceed the
160\,ms budget more frequently. Nevertheless, success increases from
44.06\% to 46.94\%, supporting the effectiveness of incorporating fresh
observations through FOP.

\begin{table}[t]
    \centering
    \caption{RAVEL serial-cycle computation time on DOM. Each sample sums
    the two expert updates and VLM encoding, excluding compensation sleeps.
    Only cycles with all three stages completed are included; startup and
    episode-tail cycles are excluded. The last column gives the percentage
    exceeding the 160\,ms computation budget.}
    \label{tab:dom-serial-cycle}
    \begin{tabular*}{\linewidth}{@{\extracolsep{\fill}}lrrr@{}}
        \toprule
        Configuration & P50 (ms) & P95 (ms) & Overrun (\%) \\
        \midrule
        Without FOP & 115.43 & 149.94 & 2.52 \\
        With FOP & 133.17 & 173.89 & 11.39 \\
        \bottomrule
    \end{tabular*}
\end{table}

\paragraph{VLASH.}
Following VLASH's training framework, we implement adaptation training
for DynamicVLA in our local codebase, fine-tuning the VLM backbone and
action expert.
Training uses global
batch size 128, with learning rates of $5\times10^{-5}$ for the
action expert and $10^{-6}$ for the VLM backbone. Training pairs the original images and language with a future-state
anchor and a demonstration action chunk shifted by a sampled delay of up
to ten control steps. The anchor is the preceding demonstration action's
pose, or the observed state at zero delay. We express the shifted pose
targets relative to this anchor and optimize the flow-matching loss over
the 20-action chunk, keeping gripper targets absolute. We select the best-performing checkpoint among
10K, 20K, and 30K steps, which is the 20K-step checkpoint. At inference, VLASH uses ten denoising steps
per 20-action chunk and requests a new chunk when ten actions remain,
giving a 400\,ms overlap window. It estimates the handoff state from the
predecessor's absolute action target and decodes new relative actions
against that state, with action quantization ratio one.
Predicting relative actions may make this adaptation sensitive to errors
in the estimated handoff state: during training, both the conditioning
state and the relative action targets depend on the chosen reference state.
At inference, the estimated state conditions the prediction and anchors
its decoding, but may differ substantially from the state actually reached
at execution. This mismatch may contribute to the poor adaptation results.

\paragraph{FASTER.}
Following FASTER's training framework, we implement adaptation training
for DynamicVLA in our local codebase, fine-tuning the VLM backbone and
action expert. Training
uses global batch size 128, with learning rates of
$2.5\times10^{-5}$ for the action expert and $10^{-6}$ for the VLM backbone.
We mix constant and horizon-aware denoising schedules with probability
$0.5$, using $\alpha=0.6$, $u_0=0.9$, and a maximum training delay of ten
steps, and evaluate the 30K-step checkpoint. At inference, FASTER
uses ten denoising steps over a 20-action horizon, an assumed delay of
four control steps (160\,ms), and an execution horizon of 12 actions
(480\,ms). The same $\alpha$ and $u_0$ are retained, and inference runs
asynchronously without compilation.
Using the baseline constant-schedule, direct full-chunk inference with
the same checkpoint yields 28.56\% success under time compensation,
comparable to the baseline DynamicVLA model and higher than the 23.78\%
achieved with FASTER inference. This indicates that the checkpoint retains
task capability. We qualitatively observed abrupt
changes in FASTER's output actions. This may be because using fewer denoising
steps for near-term actions is less well suited to predicting relative
actions, leading to these discontinuities.

\FloatBarrier
\subsection{Effect of Rolling Refinement}
\label{app:dom-full-chunk}
We investigate the benefit of rolling refinement over generating complete
action chunks for dynamic control. Specifically, we examine whether access
to updated observations during denoising is sufficient when actions are
still generated as complete chunks. We evaluate full-chunk denoising with
temporal-offset conditioning, both with and without FOP.

The action expert is initialized from the released DynamicVLA checkpoint
and trained on shifted observation--action pairs for 100K steps, using a
shared noise level across each chunk and explicit temporal-offset conditioning.
At inference, CI \& LAAS schedules 20-action chunks generated with ten
denoising steps and a fixed eight-step temporal offset. The FOP variant
starts from this trained expert and trains the FOP adapters for 100K steps.
During full-chunk denoising, FOP encodes the latest image and robot state at
steps 1 and 6. The expert uses the resulting features through cross-attention
at every denoising step, reusing them until the next refresh.
Both variants are evaluated on 90 DOM scenes with 20 trials per scene.

\begin{table}[t]
    \centering
    \caption{Full-chunk denoising on DOM, with 1,800 episodes per setting.
    Elapsed control steps reports the median number of control steps elapsed during chunk generation.}
    \label{tab:dom-full-chunk}
    \begin{tabular*}{\linewidth}{@{\extracolsep{\fill}}lcc@{}}
        \toprule
        Configuration & Success rate & Elapsed control steps \\
        \midrule
        Full-chunk & 16.78\% & 9 \\
        Full-chunk + FOP & 21.94\% & 11 \\
        \bottomrule
    \end{tabular*}
\end{table}

With temporal alignment at inference, the trained full-chunk policy achieves
16.78\% success, compared with 35.50\% for the original CI \& LAAS baseline.
Adding FOP raises success to 21.94\% (Table~\ref{tab:dom-full-chunk}).

A full-chunk policy completes all ten denoising steps before releasing
new actions. Across the completed evaluation episodes, the logged inference
time, excluding sleeps used to align simulation timing, has a P50/P95 of
370/458\,ms without FOP and 449/525\,ms with FOP.
Generation spans a median of nine and eleven control steps, respectively.
Although inference uses temporal alignment, new actions become available
only after the full chunk has been generated. FOP improves success by
incorporating fresh observations, while the wait for full-chunk generation
continues to limit response speed.

These results suggest that temporally shifted training and temporal-offset
conditioning alone are insufficient to improve full-chunk control.
In \method{}, streaming-aligned training works together with rolling
refinement: partially denoised future actions persist across updates,
allowing each update to incorporate updated context and fresh observations,
refine the buffer, and release its head actions. This progressively corrects
future actions before execution and reduces the wait for new actions
compared with generating a complete chunk from scratch.

\FloatBarrier
\section{Real-World Evaluation Details}
\label{app:real-world-details}

All three tasks use a seven-degree-of-freedom UFACTORY xArm~7 robot
and two RealSense cameras. Images are captured at 30\,Hz, and the robot
operates at a nominal control rate of 30\,Hz. We collect demonstrations
using GELLO~\citep{wu2023gello}.
Policy inference runs on a workstation with an NVIDIA GeForce RTX~5090
GPU, using Ubuntu~22.04, PyTorch~2.7.1, and CUDA~12.8.

\begin{figure}[t]
    \centering
    \includegraphics[width=\linewidth]{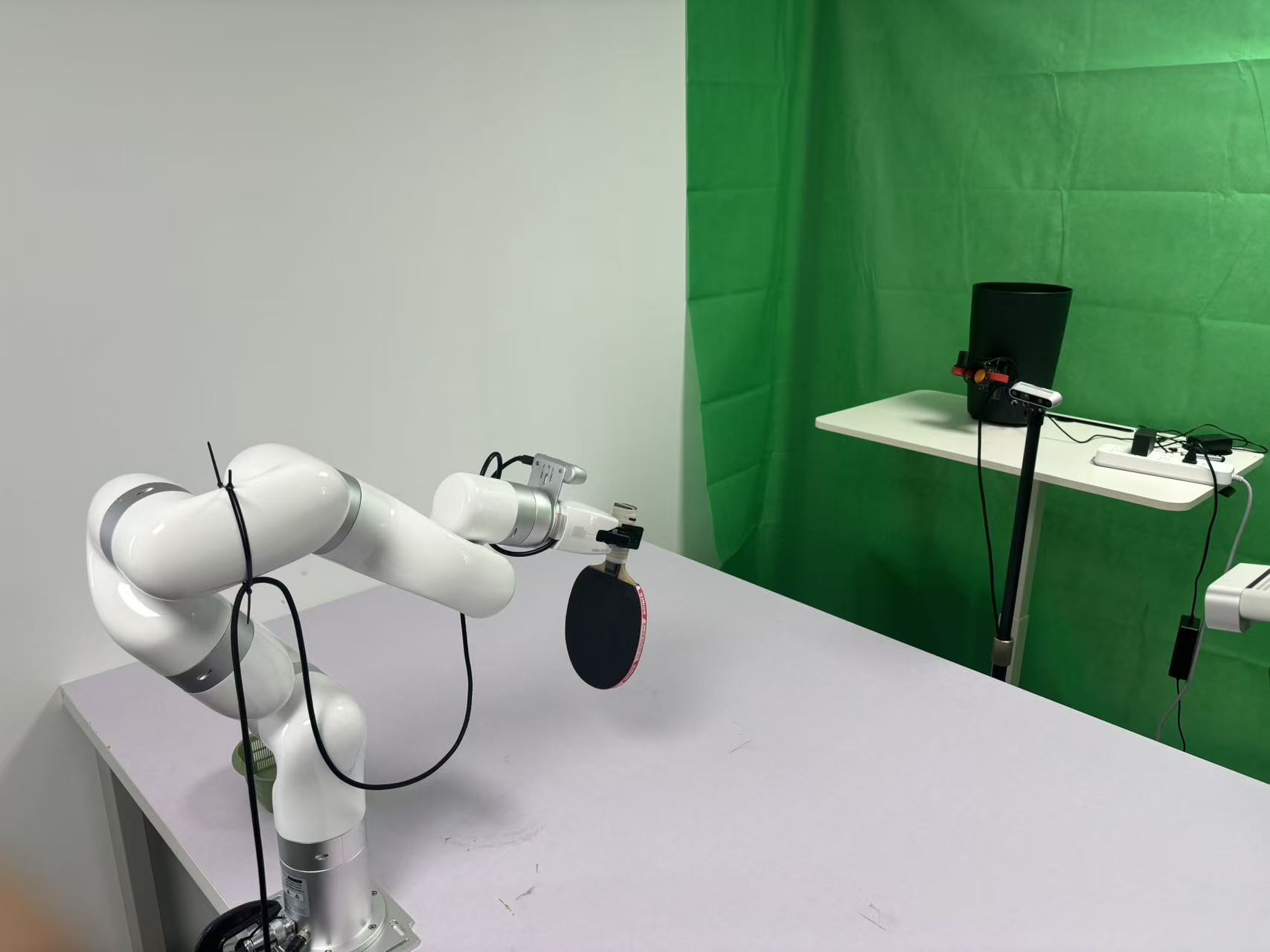}
    \caption{Real-world table-tennis return setup, with an xArm7 robot equipped with a paddle and a ball launcher.}
    \label{fig:real-world-setup}
\end{figure}

\subsection{Task Setups}
\label{app:real-task-setups}

For table-tennis return and conveyor-belt grasping,
we compare Sync, Naive Async, RTC, VLASH, and \method{}. Each method
uses the same demonstration dataset within a task, with inference
parameters selected separately for each method.

\paragraph{Table-Tennis Return.}
\label{app:table-tennis-return}

The robot uses a paddle to intercept and return an incoming ball
(Figure~\ref{fig:real-world-setup}). A ball launcher serves the incoming
balls to provide consistent conditions across methods. The policies use
joint-state feedback to predict joint-position targets.
Successful execution requires the paddle to reach the ball's path and
swing at the appropriate time.
We collect 299 teleoperated trajectories for training.

\paragraph{Conveyor-Belt Grasping.}
\label{app:conveyor-belt-grasping}

A conveyor belt moves a cube through the robot's workspace. The robot
must align the gripper with the moving cube, close the gripper, and lift
the cube from the belt. Inference latency can cause the cube to pass the
grasp point before the gripper closes.
We collect 272 teleoperated trajectories for training.

\paragraph{Tennis-Ball Grasping.}

We compare Sync, Naive Async, RTC, and \method{} over 20 trials per method,
with a 30-s timeout. Success requires grasping the ball and placing it in
the basket.

We collect 66 teleoperated demonstrations at 30\,Hz (10.4 minutes),
using 64 for training and two for validation. The policies
use two camera views and predict joint targets and gripper commands.

\subsection{Training Configurations}
\label{app:real-training}

\paragraph{Table-Tennis Return and Conveyor-Belt Grasping.}
Table-tennis return and conveyor-belt grasping follow the same training
procedure, with a separate policy trained on each task's demonstrations.
For each task, we first fine-tune pretrained $\pi_{0.5}$ using rank-64
LoRA adapters in the action expert, together with trainable action and
flow-timestep projections, for 10K steps to obtain the task policy.
Sync, Naive Async, and RTC use this checkpoint directly, without further
training; they differ in their inference and execution procedures.
VLASH and \method{} both start from this task policy and train for an
additional 10K steps on the same task data. VLASH updates the full model,
whereas \method{} trains the action expert with the VLM backbone frozen.
For \method{}, streaming-aligned training pairs earlier observations
with action targets for later control steps and conditions the action
expert on the corresponding temporal offset. A position-dependent noise
schedule matches the partially denoised action buffers encountered during
rolling inference. Table~\ref{tab:dynamic-real-training-config}
summarizes the training settings used for both tasks.

\begin{table}[!t]
    \centering
    \caption{Training configurations shared by table-tennis return and
    conveyor-belt grasping. Each task uses its own task policy and
    demonstration dataset. Steps for VLASH and \method{} denote
    additional training after task-policy fine-tuning. Batch size is global.}
    \label{tab:dynamic-real-training-config}
    \small
    \setlength{\tabcolsep}{2pt}
    \begin{tabular*}{\columnwidth}{@{\extracolsep{\fill}}lccc@{}}
        \toprule
Setting & Task policy & VLASH & \method{} \\
        \midrule
Initialization & $\pi_{0.5}$ & \multicolumn{2}{c}{Task policy} \\
Training scope & LoRA & Full model & Action expert \\
Trainable parameters & 29.89M & 3.35B & 431.15M \\
Training steps & 10K & 10K & 10K \\
Batch size & 128 & 128 & 64 \\
Learning rate & $5\times10^{-5}$ & $5\times10^{-5}$ & $5\times10^{-5}$ \\
LR decay steps & 20K & 10K & 10K \\
Warmup steps & 500 & 500 & 250 \\
        \bottomrule
    \end{tabular*}
\end{table}

\paragraph{Tennis-Ball Grasping.}
The task policy is fine-tuned from pretrained $\pi_{0.5}$.
Sync, Naive Async, and RTC use its 10K-step checkpoint directly;
\method{} starts from the same checkpoint and trains for a further 5K steps.
Table~\ref{tab:grasp-training-config} summarizes the training configurations.

\begin{table}[!t]
    \caption{Training configurations for tennis-ball grasping.}
    \label{tab:grasp-training-config}
    \begin{center}
    \begin{tabular*}{\columnwidth}{@{\extracolsep{\fill}}lcc@{}}
        \toprule
Setting & Task policy & \method{} \\
        \midrule
Initialization & $\pi_{0.5}$ & Task policy \\
Training scope & LoRA & Action expert \\
Trainable parameters & 29.89M & 431.15M \\
Training steps & 10K & 5K \\
Batch size & 128 & 128 \\
Learning rate & $5\times10^{-5}$ & $10^{-5}$ \\
LR decay steps & 10K & 5K \\
Warmup steps & 500 & 250 \\
        \bottomrule
    \end{tabular*}
    \end{center}
    \par
    \begin{minipage}{\columnwidth}
    All configurations use AdamW, weight decay $10^{-10}$,
    and gradient clipping norm 1.0. Batch size is global.
    \end{minipage}
\end{table}

\subsection{Evaluation Metrics}
\label{app:real-metrics}

\paragraph{Observation age.}
For both dynamic tasks, observation age measures the elapsed time from
the policy's receipt of the source image used to generate an action to
the dispatch of that action to the robot. We report the median (P50)
and 95th percentile (P95) in milliseconds. The measurement ends at
command dispatch and excludes the subsequent physical execution time.
Success rates are computed over trials, while observation age describes
the freshness of visual information at action dispatch.

\paragraph{High-frequency residual power.}
Inspired by the end-effector spectral-power analysis of
\citet{sojib2026efficientmetric}, we define $E_{\mathrm{HF}}$ using linear
detrending, length normalization, and the 1--15\,Hz frequency band.
We resample TCP positions at 30\,Hz and divide each trajectory into
non-overlapping 3-s windows. Within each window, we subtract a fitted
linear trend from each position coordinate and sum the remaining spectral
power at frequencies of 1\,Hz and above:
\begin{equation}
    E_{\mathrm{HF}}
    = \sum_{d \in \{x,y,z\}} \sum_{1\,\mathrm{Hz} \leq f \leq 15\,\mathrm{Hz}}
      \frac{c_f}{N^2}\left|\mathcal{F}\!\left[
      x_d(t) - (a_dt+b_d)\right](f)\right|^2,
\end{equation}
where $x_d(t)$ is the TCP position along axis $d$, $a_dt+b_d$ is its fitted
trend, and $N$ is the number of samples in the window.
$\mathcal{F}$ denotes the unnormalized discrete Fourier transform.
Within the selected band, $c_f=2$ for $1\leq f<15$\,Hz and
$c_f=1$ at the 15\,Hz Nyquist frequency.
We take the median window score for each trajectory.

\paragraph{TCP jerk.}
We resample measured Cartesian velocity at 30\,Hz and
estimate its second derivative using a local cubic fit:
\begin{equation}
    J_{\mathrm{RMS}} = \sqrt{\frac{1}{M}\sum_{i=1}^{M}
    \left\|\mathbf{j}(t_i)\right\|_2^2},
    \qquad \mathbf{j}(t_i) =
    \left.\frac{\mathrm{d}^2\mathbf{v}(t)}{\mathrm{d}t^2}\right|_{t=t_i},
\end{equation}
where $\mathbf{v}(t)$ is the measured Cartesian TCP velocity,
$\mathbf{j}(t_i)$ is the estimated jerk vector, and $M$ is the number of
valid samples in the analysis interval. We compute both metrics over each
complete recorded trajectory, including successful and failed attempts.
Table~\ref{tab:real-grasping} reports the median of the trajectory-level scores
for each method.

Both smoothness metrics depend on motion speed and range: slower or smaller
movements can yield lower scores. They should therefore be interpreted
alongside task success.

\end{document}